\documentclass{sigchi-ext}
\usepackage[T1]{fontenc}
\usepackage{textcomp}
\usepackage[scaled=.92]{helvet} 
\usepackage{graphicx} 
\usepackage{balance}  
\usepackage{booktabs} 
\usepackage{ccicons}  
\usepackage{ragged2e} 

\usepackage{xspace}

\usepackage{graphicx,subfigure}
\usepackage{ctable}
\usepackage{comment}
\usepackage{multicol}
\usepackage{multirow}
\usepackage{epsfig}
\usepackage{pifont}

\usepackage{array}
\usepackage{multirow}

\usepackage{hyperref}

\def\plaintitle{SIGCHI Extended Abstracts Sample File: Note Initial
  Caps} 
 \def\emptyauthor{}
 \def\plainkeywords{Authors' choice; of terms; separated; by
   semicolons; include commas, within terms only; required.}

\title{Explaining the Road Not Taken}

\newcommand{\hua}[1]{{\small\textcolor{orange}{\bf [#1 --Hua]}}}

\newcommand{\system}{\emph{Format Probers}\xspace}
\newcommand{\xaiqb}{XAI Question Bank\xspace}

\newcommand{\mint}{model interpretation\xspace}
\newcommand{\mints}{model interpretations\xspace}

\newcommand{\inter}{interpretation\xspace}
\newcommand{\inters}{interpretations\xspace}

\newcommand{\Inters}{Interpretations\xspace}

\newcommand{\dnns}{deep neural networks\xspace}

\newcommand{\dnn}{deep neural network\xspace}

\numberofauthors{2}
\author{%
  \alignauthor{%
    \textbf{Hua Shen}\\
    \affaddr{The Pennsylvania State University} \\
    \affaddr{University Park, PA 16802, USA} \\
    \email{huashen218@psu.edu} }
    \vfil \alignauthor{%
    \textbf{Ting-Hao (Kenneth) Huang}\\
    \affaddr{The Pennsylvania State University}\\
    \affaddr{University Park, PA 16802, USA}\\
    \email{txh710@psu.edu} } 
    }

\definecolor{linkColor}{RGB}{6,125,233}
\hypersetup{%
  pdftitle={\plaintitle},
  pdfauthor={\emptyauthor},
  pdfkeywords={\plainkeywords},
  bookmarksnumbered,
  pdfstartview={FitH},
  colorlinks,
  citecolor=black,
  filecolor=black,
  linkcolor=black,
  urlcolor=linkColor,
  breaklinks=true,
}

\begin{document}

\CopyrightYear{2021}
\setcopyright{rightsretained}
\conferenceinfo{ACM CHI Workshop on Operationalizing Human-Centered Perspectives in Explainable AI,}{May 8--9, 2021, Online Virtual Conference (originally Yokohama, Japan)}
\isbn{}
\doi{}
\copyrightinfo{\acmcopyright}

\maketitle

\RaggedRight{}

\begin{abstract}

It is unclear if existing \inters of \dnn models respond effectively to the needs of users.
This paper summarizes the common \textit{forms} of explanations (such as feature attribution, decision rules, or probes) used in over 200 recent papers about natural language processing (NLP), and compares them against user questions collected in the \xaiqb~\cite{liao2020questioning}.
We found that although users are interested in explanations for \textit{the road not taken}~--- namely, why the model chose one result and not a well-defined, seemly similar legitimate counterpart~--- most \mints cannot answer these questions.

\end{abstract}

\ccsdesc[500]{Human-centered computing~Usability testing}
\ccsdesc[500]{Human-centered computing~User studies}





%
%




\section{Introduction}
\label{sec:introduction}

Researchers have attempted to produce \mints for \dnns~\cite{montavon2018methods} under the broader umbrella of Explainable Artificial Intelligence (XAI).
The primary objective of this line of research is two-fold~\cite{jacovi-goldberg-2020-towards}:
to create \inters that faithfully characterize the models' behavior ({\em i.e.}, are \textit{faithful}), 
and to improve user trust or understanding of black-box algorithms ({\em i.e.}, appear \textit{plausible}). 
However, this objective does not always align with the practical needs of users.
Recent studies reveal that a faithful or plausible \mint can still be useless, or even harmful, to its users. 
For example, our previous work found that showing users visual explanations (saliency maps) decreased~--- \textit{not} increased~--- users' ability to make sense of the mistakes made by neural image classifiers~\cite{shen2020useful}.
Another study showed that visual explanations may not alter human accuracy or trust in the model~\cite{chu2020visual}.
Recent work in XAI has begun to mitigate this misalignment~\cite{eiband2018bringing}; one example is collecting algorithm-informed user demands from real-world practices~\cite{liao2020questioning}.

This paper takes a closer look into the gap between user need and current XAI.
Specifically, we survey the common \textit{forms} of explanations, such as feature attribution~\cite{chen-ji-2020-learning,lei-etal-2016-rationalizing}, decision rule~\cite{saha-etal-2020-prover,jiang:2019:emnlp}, or probe~\cite{lin-etal-2019-open,ettinger2020bert}, used in 218 recent NLP papers, and compare them to the 43 questions collected in the \xaiqb~\cite{liao2020questioning}.
We use the forms of the explanations to gauge the misalignment between user questions and current NLP explanations.

\marginpar{%
  \vspace{-165pt} 
  \fbox{%
    \begin{minipage}{0.98\marginparwidth}
      \textbf{Explainable AI Formats-I} \\
      
      \vspace{0.6pc} 
      \textbf{1-Feature Attribution (FAT) [43.99\%]
      :} highlight the sub-sequences in input texts~\cite{chen-ji-2020-learning,lei-etal-2016-rationalizing}, Typical question~\cite{mudrakarta-etal-2018-model}:
      \begin{itemize}
        \vspace{-0.8em}
          \item {\em How can we attribute the systems' predictions to input features?}
      \end{itemize}

      \vspace{-0.3em} 
      \textbf{2-Tuple/Graph (TUP) [10.15\%]:} explain model reasoning process with tuples/ trees/ graphs~\cite{stadelmaier2019modeling,moon-etal-2019-opendialkg}. Typical question~\cite{chen2019multihop}:
      \begin{itemize}
        \vspace{-0.8em}
          \item {\em How does the system use reasoning graphs to arrive at the answer?} 
      \end{itemize}
      
      \vspace{-0.3em}  
      \textbf{3-Concept/Sense (CPT) [9.72\%]:} convert to human interpretable concepts or terminologies~\cite{chang2019does,schwarzenberg2019neural}. Typical question~\cite{panigrahi-etal-2019-word2sense}:
      \begin{itemize}
        \vspace{-0.8em}
          \item {\em What sense does the system's intermediate representation make?} 
      \end{itemize}
      
      \vspace{-0.3em} 
      \textbf{4-Rule/Grammar (RUL) [9.61\%]:} extract executable rules or logic for model decisions.~\cite{jiang:2019:emnlp,pezeshkpour-etal-2019-investigating}. Typical question~\cite{ribeiro2018anchors}:
      \begin{itemize}
        \vspace{-0.8em}
          \item {\em How can we explain the system's behavior with executable rules?} 
      \end{itemize}
      
    \end{minipage}}
    \label{sec:sidebar1} 
}

\section{Gauging Explainable AI Gaps Using Forms}
\label{sec:study}

\begin{figure*}[ht]
    \centering
    \includegraphics[width=1.0\textwidth]{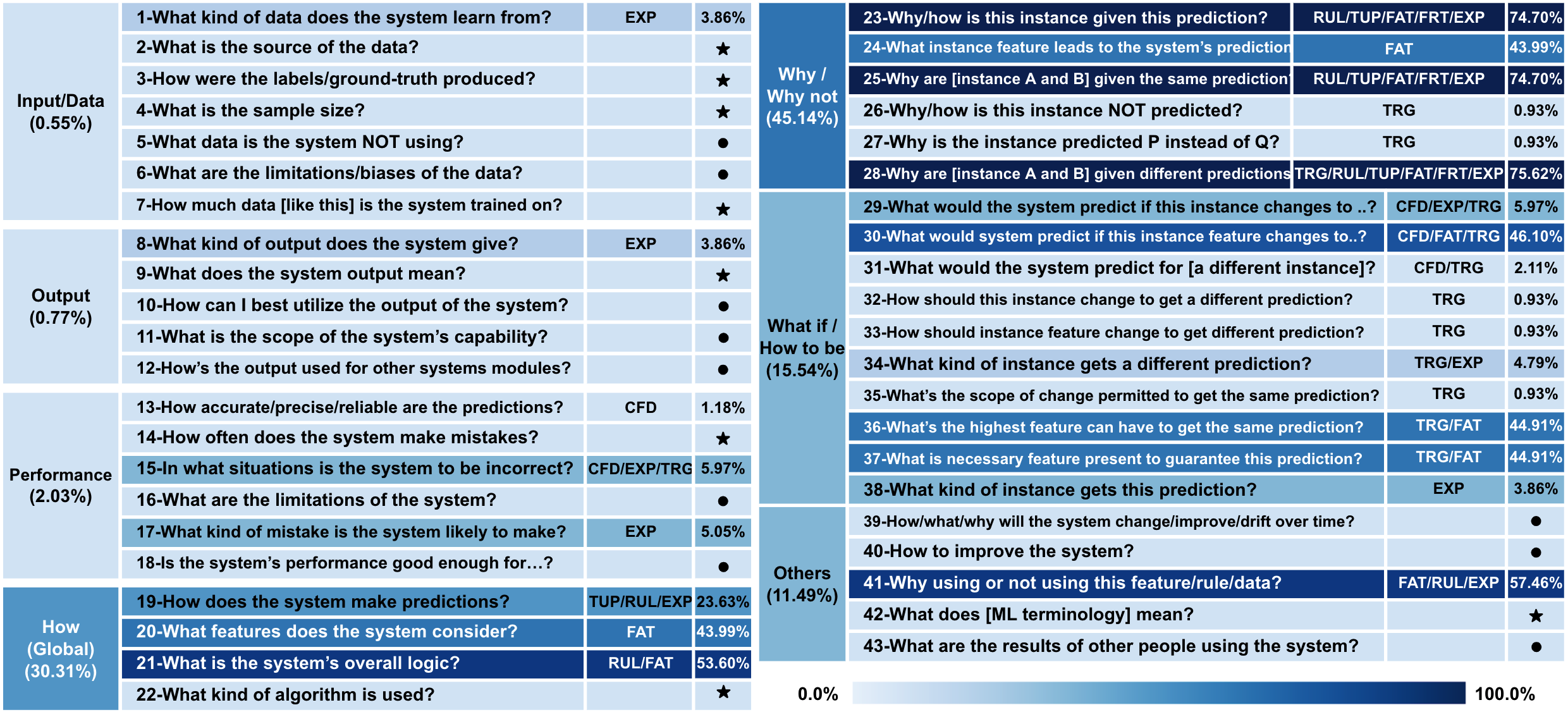}
    \caption{The questions in \xaiqb, heat-mapped by the estimated percentage (\%) of NLP XAI studies attempting to answer them. 
    ($\bullet$: questions that can \textit{not} be answered by most NLP XAI studies; $\star$: questions that can likely be answered by the AI system's meta information.)}
    \label{fig:xai_question_bank}
 \end{figure*}

Liao {\em et. al}~\cite{liao2020questioning} developed the \xaiqb, a set of prototypical questions users might ask about AI systems.
This paper investigates how well these questions are answered by current XAI work in NLP.
We collected 218 recent NLP papers about interpretability, analyzed the \textit{forms} of interpretations these papers researched ({\em e.g.}, feature attribution, decision rules, etc.), and used these forms to associate each paper to the questions it tried to answer.
This section overviews our two-step procedure.

\paragraph{Step 1: Survey the Forms of \Inters in NLP Papers.}
We first reviewed 218 explanation studies published in the NLP field between 2015 and 2020, and 
came up with 12 common XAI forms.
We defined a paper as an NLP explanation study if: 
{\em (i)} its motivation was to explain or analyze NLP models, tasks, or datasets; or 
{\em (ii)} it aimed to develop more explainable NLP models, tasks, or datasets; or 
{\em (iii)} the explanation format is natural language.
Given those definitions, we decided on a set of search keywords ({\em e.g.}, ``explain'', ``interpretation''),
a list of top-tier publications and conference proceedings ({\em e.g.}, ACL, EMNLP),
and a range of publication years.
Within the venues and years, we collected all papers whose titles or abstracts contained those keywords.
Then we read each paper and added the related papers that it cited about ``NLP explanation'' 
into our collections.
Our ultimate list of papers covered various conferences, workshops, and other research fields ({\em e.g.}, human-computer interaction).

%
%

Our definition of ``interpretation form'' is \textit{how the study represents its explanation results}.
In this paper, we present 12 different interpretation forms. 
We started with four commonly used forms, including ``feature attribution~\cite{chen-ji-2020-learning,lei-etal-2016-rationalizing},'' ``tuple/graph~\cite{moon-etal-2019-opendialkg,stadelmaier2019modeling},'' ``free text~\cite{lee-etal-2020-lean,explainyourself},'' and ``example~\cite{han2020explaining,yeh2018representer}.''
Then we read each paper, assigned a form to it, and added new forms into our scheme as needed.

We present the 12 forms with their abbreviations, format weight, brief definitions, representative work, and one typical question in sidebars on page 2 to 4. 
We released our data\footnote{Please see details at \url{https://human-centered-exnlp.github.io}.}, which contains the list of the 218 NLP explanation papers with each paper's title, year, venue, and form annotations.

We then computed what percent of the 218 XAI papers used each type of interpretation form ({\em i.e.,} format weight).
We gave each paper a weight of $1$.
If the paper used only one form type, we assigned $1$ to the form.
If the paper used multiple interpretation forms, we assigned all its applicable forms an equal weight totaling $1$. 
To obtain the final percentage of each form type, we added up its scores among all papers and divided by the count of papers.

As shown in the sidebars, the most common form of current NLP explanations~--- around 44\% of related studies~--- is to highlight features ({\em e.g.}, tokens or sentences) within input text.
%
%
\marginpar{%
  \vspace{-335pt} 
  \fbox{%
    \begin{minipage}{0.98\marginparwidth}
      \textbf{Explainable AI Formats-II} \\
      
      \vspace{0.6pc} 
      \textbf{5-Probing (PRB) [7.79\%]:} classify representation with specific diagnostic dataset~\cite{lin-etal-2019-open,ettinger2020bert}. Typical question~\cite{hewitt2019designing}:
      \begin{itemize}
        \vspace{-0.8em}
          \item {\em What linguistic properties does the system's representation have?} 
      \end{itemize}
      
      \vspace{-0.3em} 
      \textbf{6-Free Text (FRT) [7.09\%]:} use natural language to explain model behavior~\cite{lee-etal-2020-lean,explainyourself}. Typical question~\cite{NEURIPS2018_4c7a167b}:
      \begin{itemize}
        \vspace{-0.8em}
          \item {\em How can we explain a system's decision using natural language justification?} 
      \end{itemize}
      
      \vspace{-0.3em} 
      \textbf{7-Example (EXP) [3.86\%]:} find most responsible training samples as explanations~\cite{han2020explaining,yeh2018representer}. Typical question~\cite{influencefunction}:
      \begin{itemize}
        \vspace{-0.8em}
          \item {\em How can we trace the system's prediction back to the training sample(s) most responsible for it?} 
      \end{itemize}
      
      \vspace{-0.3em} 
      \textbf{8-Projection Space (PSP) [3.82\%]:} project dense vectors into low-dimensional space~\cite{s2svis:2018:ieee,wu2019self}. Typical question~\cite{aubakirova-bansal-2016-interpreting}:
      \begin{itemize}
        \vspace{-0.8em}
          \item {\em How can we project the system's high-dimensional representation to a human-understandable space?} 
      \end{itemize}
      
    \end{minipage}}
    \label{sec:sidebar2} 
}
Approximately 10\% of NLP explanation work leverages a tuple, rule, or concept format to demonstrate the model's reasoning process.
Other studies use a probe to diagnose what information the model representation can embed, or directly explains model behavior with free text.
Less than 5\% of algorithms use training data examples, projection space,
or output confidence scores to visualize NLP explanations. 
Fewer than 5 papers explain NLP models with word cloud, trigger, or image formats.

\paragraph{Step 2: Compare Against User Questions in the \xaiqb.}
The \xaiqb collected user questions for AI explanations from real-world user needs~\cite{liao2020questioning}.
It consists of 43 questions within 7 categories about AI systems, as detailed in Figure~\ref{fig:xai_question_bank}.
The prototypical questions are identified by analyzing current XAI algorithms and interviewing UX and design practitioners in IBM product lines.
We annotated each user question in the \xaiqb with all applicable forms identified in Step 1.
The principle we used to annotate a user question with forms was \textit{if the format had ever answered similar questions among our collected studies}.
Specifically, in the first step, we noted typical questions the form answered in the literature exemplified in the sidebars.
For instance, the ``example'' form primarily answers ``What are the training instances most responsible to support this prediction?''~\cite{han2020explaining,influencefunction}
Then we inspected each question in the \xaiqb and looked for similar questions we collected for the 12 forms.
We labeled the user question with its corresponding form when the form was used to answer similar questions in the literature.
For instance, to answer the user question ``How does the system make a prediction?'' we can explain AI systems to users using executable logic rules ({\em i.e.}, RUL), decision-reasoning graphs ({\em i.e.}, TUP), or by showing each class's representative examples ({\em i.e.}, EXP).
Afterwards, we calculated each user question's weight by adding all its labeled formats' weights.
The user question weight roughly approximates the proportion of published NLP papers that can answer this question.
This resulted in the weighted \xaiqb as shown in Figure~\ref{fig:xai_question_bank}, which provides intuitive visualization of the NLP research's attention to answering user questions.
Note that XAI forms may evolve rapidly due to proliferation of XAI studies, but we can extend the collected XAI forms and repeat the gap-gauging process easily.

\section{The Need to Explain the Road Not Taken}
\label{sec:result}

%
\marginpar{%
  \vspace{-300pt} 
  \fbox{%
    \begin{minipage}{1.0\marginparwidth}
      \textbf{Explainable AI Formats-III
      }\\
     
      \vspace{0.6pc} 
      \textbf{9-Confidence Score (CFD) [1.18\%]:} leverage model prediction probability to show confidence~\cite{hase2020evaluating,exbert:2020:acl}. Typical question~\cite{feng2019can}:
      \begin{itemize}
        \vspace{-0.8em}
          \item {\em How much uncertainty does the system have on its prediction?} 
      \end{itemize}
      
      \vspace{-0.3em} 
      \textbf{10-Word Cloud (WCL) [1.16\%]:} generate word cloud using model representations~\cite{pappas2014explaining,chen-ji-2020-learning}. Typical question~\cite{lertvittayakumjorn2020find}:
      \begin{itemize}
        \vspace{-0.8em}
          \item {\em What are the input patterns that activate the system prediction?} 
      \end{itemize}
      
      \vspace{-0.3em} 
      \textbf{11-Trigger (TRG) [0.93\%]:} make change to trigger models to produce counterfactual predictions~\cite{feng-etal-2018-pathologies,sankar-etal-2019-neural}. Typical question~\cite{wallace2019universal}:
      \begin{itemize}
        \vspace{-0.8em}
          \item {\em What are the token sequences that trigger a model to produce a different prediction?} 
      \end{itemize}
      
      \vspace{-0.3em} 
      \textbf{12-Images (IMG) [0.70\%]:} visualize model representations by token-related images~\cite{panchenko2017unsupervised}. Typical question~\cite{tan2020vokenization}:
      \begin{itemize}
        \vspace{-0.8em}
          \item {\em How to map the system's language tokens to their related images?} 
      \end{itemize}
    \end{minipage}}
    \label{sec:sidebar3} 
}

While 9 out of 43 questions in the \xaiqb are about how AI systems \textbf{can} provide specific predictions ({\em i.e.}, Q19-21,23-25,36-38), 16 questions are about what AI systems \textbf{cannot} achieve and why ({\em e.g.}, Q5-6,11,15-16,26- 35,41).
Many of these under-answered or unanswered questions are \textit{counterfactual} questions, such as ``Why did the model predict P instead of Q for this instance?''
These questions can probably be answered by a trigger (TRG), but only three papers out of the surveyed 218 focused on counterfactual explanations~\cite{wallace2019universal,feng-etal-2018-pathologies,sankar-etal-2019-neural}.
Furthermore, we speculate that many of these questions assume one or more well-defined, seemly similar legitimate counterpart labels ({\em e.g.}, \textit{positive} versus \textit{negative}, \textit{dog} versus \textit{cat}), in which the user wonders why the system choose one over the other.
More fundamentally, the fact that users want to know both \textit{why} and \textit{why not} the AI system made certain predictions may suggest that users' goals are often to \textbf{gain a \textit{global view} of how the AI system works}.

It is worth noting that more NLP work has begun to generate counterfactual examples ({\em i.e.}, ``contrastive sets''), often with the purpose of learning robust NLP models~\cite{gardner-etal-2020-evaluating,Kaushik2020Learning,wu2021polyjuice}.
These methods could be extended to generating counterfactual explanations. 
As counterfactual explanations have been explored in other domains, such as computer vision~\cite{chang2018explaining}, tabular data classification~\cite{dice:2020:fat}, and interactive tools~\cite{vice:iui:2020}, recent NLP work has begun to focus more attention on developing counterfactual explanations~\cite{jacovi2021contrastive,ross2020explaining}\footnote{We did not include these recent studies in our paper collection because they were published after our paper-collecting and analysis process.}.

\paragraph{Which Road Do You Want Explanations For?}
Developing counterfactual explanations in NLP can be challenging.
It is not always easy to tell \textbf{which counterfactual predictions} should be explained.
Jacovi {\em et al.} submitted a good example~\cite{jacovi2021contrastive}:
When people ask ``Why did the AI system choose to hire Person X?'' they could mean either 
``Why did the AI system choose to hire Person X rather than not hire Person X?'' or 
``Why did the AI system choose to hire Person X rather than hire Person Y?''
Liao {\em et al.} suggested that AI explanations can be provided in an \textit{interactive} manner, allowing people to ``explicitly reference the contrastive outcome and ask follow-up \emph{what if} questions''~\cite{liao2020questioning}.
As ambiguous and underspecified language can be common, more research is required to help users spot the meaningful counterfactual predictions they actually care about.



\section{Discussion}

\paragraph{User Questions Beyond the Scope of the Current XAI.}
In another finding included in Figure~\ref{fig:xai_question_bank}, we observed 8 questions ({\em i.e.}, labeled $\star$) that can be addressed by the \textit{meta information} in AI algorithms (such as ``What is the source of the data?'') but that XAI forms do not answer.
However, we find 10 questions ({\em i.e.}, labeled $ß\bullet$) that the XAI forms cannot address well.
These questions mainly inquire about the \textbf{limitation, potential utility, or capability scope} of AI systems ({\em e.g.}, ``What are the limitation/biases of the data?''), which are seldom introduced in XAI studies.
We posit XAI algorithm developers should use these questions to develop corresponding XAI methods or to clarify capability scope, system utility, and limitation in the methods.

\paragraph{Limitations.}
We are aware of several limitations of our work.
First, this paper focuses on NLP applications, but the \xaiqb captures user questions for a broader spectrum of AI systems.
Second, the \xaiqb provides an in-depth analysis of lay users' needs, while the user population for the NLP papers included in our study are broader, such as domain experts~\cite{feng-etal-2020-explainable,valenzuela2018lightly} and AI practitioners~\cite{lime,bastings-filippova-2020-elephant}.
Finally, using forms of interpretation to associate papers with user questions inevitably overlooks some information.
For instance, the ``probing'' form does not appear in the \xaiqb.
This could be caused by the fact that some particular forms of interpretations, such as probing methods, are primarily developed for AI practitioners rather than lay people.

\section{Conclusion}
\label{sec:conclusion}

Our analysis explicates the gaps between what users want and the current focus of XAI research in NLP.
Questions like ``Why is this instance given this prediction?'' were studied extensively, and can be answered by five different interpretation formats ({\em i.e.}, ``rule/grammar,'' ``tuple/graph,'' ``feature importance,'' ``free text,'' and ``example'').
Meanwhile, 16 out of 43 user questions in the \xaiqb are relevant to counterfactual inquiries, such as
``Why did the model predict P instead of Q for this instance?'',
but only a handful of papers have tried to produce counterfactual explanations.
We learned that users want to know the decision scope of AI systems, including what the AI system can and cannot achieve.

XAI researchers can collaborate with user-experience (UX) designers to mitigate this misalignment. 
In particular, XAI algorithm developers can produce more counterfactual explanations for answering global and local counterfactual questions, or directly generate AI explanations that can explain both \textit{can} and \textit{cannot} questions ({\em e.g.}, tree-based rules).
On the other hand, one XAI form may not be enough to satisfy practical user demands for understanding \textit{can} and \textit{cannot} questions simultaneously.
Therefore, XAI UX designers can combine multiple forms and algorithms to meet real-world user requirements.
Since awareness of new explainable AI forms can change user demand~\cite{lim2009assessing,liao2020questioning}, 
perhaps XAI researchers can leverage the variety of forms to 
to respond more effectively to real-world user needs.

\balance{} 

\bibliographystyle{SIGCHI-Reference-Format}
\bibliography{sample.bib}

\end{document}